# Deep Confidence: A Computationally Efficient Framework for Calculating Reliable Errors for Deep Neural Networks

Isidro Cortés-Ciriano[1,*] and Andreas Bender[1]

[1]Centre for Molecular Informatics, Department of Chemistry, University of Cambridge, Lensfield Road, Cambridge, CB2 1EW, United Kingdom.

[*]Corresponding author: isidrolauscher@gmail.com

## Abstract

Deep learning architectures have proved versatile in a number of drug discovery applications, including the modelling of *in vitro* compound activity. While controlling for prediction confidence is essential to increase the trust, interpretability and usefulness of virtual screening models in drug discovery, techniques to estimate the reliability of the predictions generated with deep learning networks remain largely underexplored. Here, we present Deep Confidence, a framework to compute valid and efficient confidence intervals for individual predictions using the deep learning technique *Snapshot Ensembling* and conformal prediction. Specifically, Deep Confidence generates an ensemble of deep neural networks by recording the network parameters throughout the local minima visited during the optimization phase of a single neural network. This approach serves to derive a set of base learners (*i.e.,* snapshots) with comparable predictive power on average, that will however generate slightly different predictions for a given instance. The variability across base learners and the validation residuals are in turn harnessed to compute confidence intervals using the conformal prediction framework. Using a set of 24 diverse $IC_{50}$ data sets from ChEMBL 23, we show that Snapshot Ensembles perform *on par* with Random Forest (RF) and ensembles of independently trained deep neural networks. In addition, we find that the confidence regions predicted using the Deep Confidence framework span a narrower set of values. Overall, Deep Confidence represents a highly versatile error prediction framework that can be applied to any deep learning-based application at no extra computational cost.

# Introduction

To date, a plethora of deep learning applications have been proposed to aid in multiple drug discovery tasks[1–7], including *de novo* drug design[8,9], automatic molecular design[10], design of focused molecule libraries[11], multi-target bioactivity prediction[12–15], compound synergy modelling[16], as well as compound property[17], single-target bioactivity[18–21] and toxicity prediction[22], among others. Multiple studies have shown that deep neural networks (DNN) often outperform commonly used algorithms when applied to model *in vitro* bioactivity data, including Support Vector Machines (SVM) and Random Forests (RF)[21,23,24]. Importantly, the higher predictive power of deep learning has been shown for both large and small data sets[13,19,20]. Hence, DNN are becoming a standard technique to model *in vitro* compound activity[1,25].

Defining the set of molecules for which a model can generate reliable predictions has been (and still is[26]) an area of intense research because an accurate estimation of prediction errors is essential for a model to inform decision-making in drug discovery[27–30]. While diverse deep learning architectures have been developed to predict compound activity as point predictions[12,18,23], modelling prediction uncertainty for small molecules using deep learning still remains an unexplored area, and this is precisely what we address in this contribution.

Among the array of methods developed to compute the uncertainty of individual predictions, conformal prediction[31,32] has been extensively applied in drug discovery over the last few years to model compound toxicity[33], *in vitro* bioactivity data[34–38], and to guide *in silico* iterative screening campaigns.[39,40] A key advantage of conformal prediction is that each prediction is given as a confidence interval, that will be more or less stringent depending on the user-selected confidence level (CL)[31]. Moreover, these predicted confidence regions are guaranteed to be valid by a solid mathematical foundation provided that the exchangeability principle is fulfilled, which is generally assumed to be the case when modelling bioactivity data[31,32,36]. The validity of a conformal predictor means that, at a user-defined CL, the measured error rate, *i.e.,* the fraction of instances for which the true value lies outside the predicted region, will not be larger than the chosen error rate (*i.e.,* 1 - CL). For instance, at a confidence level of 80%, at least 80% of the predicted confidence regions will contain the true value. Thus, in contrast to most other prediction error methods, which do not guarantee a lower bound for the fraction of the predictions that will be correct, conformal prediction is a very powerful technique in practice;

for instance, in a drug discovery project it permits to select compounds for further experimental testing while controlling for the fraction of instances whose experimentally measured bioactivity will be above a cut-off value of interest[36,39,40].

Another major advantage of conformal prediction is that, as opposed to *e.g.,* Bayesian methods[41–45], it requires little computational cost beyond the training of the underlying algorithms, no parameter optimization, and can be coupled to any machine learning algorithm[36]. In the deep learning community, previous work on estimating predictive uncertainty for DNN used both Bayesian and non-Bayesian approaches[46–51]. In addition, post-processing of the softmax layer, although it is not a measure of confidence[52], has proved useful to deliver calibrated class probabilities in classification problems[53,54]. However, to the best of our knowledge, the integration of conformal prediction and deep learning to model bioactivity data has not been explored to date.

Here, we propose Deep Confidence, a versatile and computationally efficient framework to compute confidence intervals for individual predictions. Deep Confidence leverages the key advantages of both deep learning and conformal prediction, and can be easily integrated into any deep learning architecture. Deep Confidence uses *Snapshot Ensembles*[55] to generate an ensemble of DNN from the training of a single DNN. To generate a Snapshot Ensemble, intermediate states of a DNN during the optimization path of its parameters are taken (Figure 1). The main advantage of Snapshot Ensembles is that these snapshots are 'free', given that no extra computation is required to generate them beyond the computation time needed to train the network. The training phase is divided into cycles, throughout which the learning rate is annealed and set back to its original value at the beginning of the next cycle. At the end of a cycle, the current state of the DNN is used as a base learner in the Snapshot Ensemble provided that the DNN has (sufficiently) converged to a local minimum during that cycle. Convergence is measured using the loss on the validation set, which in the current study was the Root Mean Squared Error (RMSE) in prediction. A cut-off value related to the uncertainty of the data at hand can be used to dismiss snapshots with low predictive power before the network starts converging; for instance, 1.2 $pIC_{50}$ units, which corresponds to about two times the experimental uncertainty of public heterogeneous $IC_{50}$ data[56,57]. Therefore, while the average error in prediction might be similar across snapshots, the predictions for a given instance calculated by each snapshot will be different[55]. The fact that the snapshot predictions are correlated but not identical adds sufficient diversity to generate an ensemble with high predictive power[58]. Once the Snapshot Ensemble has been generated, the predictions of each snapshot on the validation and test sets are used to generate a conformal predictor.

We here show using 24 diverse bioactivity data sets from ChEMBL version 23 (Table 1) that Deep Confidence generates valid confidence intervals, and that these span a narrower range of values than those calculated using RF-based conformal predictors while leading to a comparable average interval size. In addition, we show that Snapshot Ensembles and ensembles of independently trained DNN lead to comparable predictive power on unseen data during the training phase. Thus, Deep Confidence represents a versatile approach to model bioactivity data using DNN ensembles at a vastly reduced computing cost while also providing an estimation of the reliability of individual predictions.

## Methods

### Data Sets

We gathered IC$_{50}$ data for 24 diverse protein targets and receptors from the ChEMBL database version 23 using the chembl_webresource_client python module[59–61]. Only IC$_{50}$ values for small molecules that satisfied the following filtering criteria were retained: (i) an activity unit equal to "nM", (ii) activity relationship equal to '=', (iii) target type equal to "SINGLE PROTEIN", and (iv) organism equal to *Homo sapiens*. IC$_{50}$ values were modeled in a logarithmic scale (pIC$_{50}$ = −log$_{10}$ IC$_{50}$). The average pIC$_{50}$ value was calculated when multiple pIC$_{50}$ values were available for the same compound. Further information about the data sets is given in Table 1 and in [62]. All data sets are provided in the Supporting Information.

**Table 1. Data sets used in this study.**

| Target preferred name | Target abbreviation | Uniprot ID | ChEMBL ID | Number of bioactivity data points |
|---|---|---|---|---|
| Alpha-2a adrenergic receptor | A2a | P08913 | CHEMBL1867 | 203 |
| Tyrosine-protein kinase ABL | ABL1 | P00519 | CHEMBL1862 | 773 |
| Acetylcholinesterase | Acetylcholinesterase | P22303 | CHEMBL220 | 3,159 |
| Serine/threonine-protein kinase Aurora-A | Aurora-A | O14965 | CHEMBL4722 | 2,125 |
| Serine/threonine-protein kinase B-raf | B-raf | P15056 | CHEMBL5145 | 1,730 |
| Cannabinoid CB1 receptor | Cannabinoid | P21554 | CHEMBL218 | 1,116 |
| Carbonic anhydrase II | Carbonic | P00918 | CHEMBL205 | 603 |
| Caspase-3 | Caspase | P42574 | CHEMBL2334 | 1,606 |
| Thrombin | Coagulation | P00734 | CHEMBL204 | 1,700 |
| Cyclooxygenase-1 | COX-1 | P23219 | CHEMBL221 | 1,343 |
| Cyclooxygenase-2 | COX-2 | P35354 | CHEMBL230 | 2,855 |
| Dihydrofolate reductase | Dihydrofolate | P00374 | CHEMBL202 | 584 |
| Dopamine D2 receptor | Dopamine | P14416 | CHEMBL217 | 479 |
| Norepinephrine transporter | Ephrin | P23975 | CHEMBL222 | 1,740 |
| Epidermal growth factor receptor erbB1 | erbB1 | P00533 | CHEMBL203 | 4,868 |
| Estrogen receptor alpha | Estrogen | P03372 | CHEMBL206 | 1,705 |
| Glucocorticoid receptor | Glucocorticoid | P04150 | CHEMBL2034 | 1,447 |
| Glycogen synthase kinase-3 beta | Glycogen | P49841 | CHEMBL262 | 1,757 |
| HERG | HERG | Q12809 | CHEMBL240 | 5,207 |
| Tyrosine-protein kinase JAK2 | JAK2 | O60674 | CHEMBL2971 | 2,655 |
| Tyrosine-protein kinase LCK | LCK | P06239 | CHEMBL258 | 1,352 |
| Monoamine oxidase A | Monoamine | P21397 | CHEMBL1951 | 1,379 |
| Mu opioid receptor | Opioid | P35372 | CHEMBL233 | 840 |
| Vanilloid receptor | Vanilloid | Q8NER1 | CHEMBL4794 | 1,923 |

**Molecular Representation**

We standardized all chemical structures to a common representation scheme using the python module *standardizer* (https://github.com/flatkinson/standardiser). Inorganic molecules were removed, and the largest fragment was kept in order to filter out counterions. We next computed circular Morgan fingerprints[63] for all compounds using RDkit (release version 2013.03.02)[64]. The radius was set to 2 and the fingerprint length to 128. In contrast to previous work[65–67], we observed that longer fingerprints (256, 512 and 1024 bits) led to marginal increases in predictive power, and hence, we used 128 bits to reduce training times.

**Machine Learning**

In the following subsections, we explain the training procedures of the 5 learning strategies explored in this study and how conformal predictors were generated for each of them. For each training strategy and data set pair 20 repeats were performed, each time randomly assigning different sets of data points to the training, validation and test sets.

- **Deep Neural Networks (DNN)**

DNN were trained using the python library Pytorch[68]. The RMSE value on the validation set was used as the loss function during the training of all DNNs reported here. We defined three hidden layers, composed of 60, 20, and 10 nodes, respectively, and used rectified linear unit (ReLU) activation. We decided to use ReLU activation following previous work where higher predictive power was obtained with ReLU as compared to the Tanh and Sigmoid functions[18]. We set the number of nodes in each layer below the number of dimensions of the input data, *i.e.* 128, to reduce the risk of overfitting[69].

The data sets were randomly split into a training set (70% of the data), a validation set (15%), and a test set (15%). For each data set, the training set was used to train a given network, whereas the validation set served to monitor the performance of the network during the training phase. The training data were processed in batches of size equal to 15% of the number of instances, as using larger batches during training decreases the generalization ability of the models[70]. The test set was used to assess the generalization capabilities of the models on unseen data during the training phase.

We used stochastic gradient descent with Nesterov momentum, which was set to 0.9 and kept constant during the training phase[71]. We allowed the networks to evolve over 3,000 epochs. To avoid overfitting we used early stopping, *i.e.,* the training of a given network was stopped if the validation loss did not decrease after 200 epochs. In addition, we used 10% dropout in the three hidden layers[13,18,72]. The DNN described in the following sections were trained using the parameter values and data partitions described above unless otherwise stated.

- **Ensembles of Deep Neural Networks**

We considered 2 strategies to generate ensembles of DNNs. Each ensemble encompassed 100 DNN as base learners, each of them trained using the same validation and training sets. The point prediction for a given instance was calculated as the mean value of the 100 individual predictions.

**Ensembles of Independently Trained Deep Neural Networks (DNN Ensemble).** The base learners in this case were 100 DNN independently trained using the same deep learning architecture. This corresponds to training the same type of DNN on the same training data, using different seed values at the start of the training phase. DNNs that failed to converge to RMSE values on the validation set smaller than 1.2 $pIC_{50}$ units were discarded and not considered in the ensemble. After reaching convergence in the training phase, each DNN was used to predict the activity for both the validation and test sets. The learning rate was initially set to 0.005, and was decreased by 40% every 200 epochs.

**Snapshot Ensembles with learning rate annealing (Snapshot Ensemble Lr Annealing).** Snapshot Ensembles were constructed by recording DNN snapshots[55] during the training phase of a single DNN. This allows to generate multiple models, each corresponding to a different local minimum, while only training one DNN[55]. Each of these snapshots served as a base learner to generate a given Snapshot Ensemble (Figure 1 and 2). Early stopping was not used in this case. The learning rate was set back to the initial value (*i.e.,* 0.005) after the completion of each cycle. During each cycle, the learning rate was annealed using step decay (the exact parameter values used are defined below). Setting back the learning rate to a high value (where high depends on the data sets being modelled) allows wider sampling of the loss landscape and to escape from local minima[73,74] (Figure 1 and 2).

In some cases, convergence was not achieved after the 3,000 epochs we used to train each network or started after few learning rate annealing cycles. We found that the loss on the validation set was either too high, indicating that no local minima were reached, or the predictions for the validation set were simply the mean $pIC_{50}$ value of the instances in the training data. Although the average error in prediction for the latter might be <1.2 pIC$_{50}$ units (*e.g.,* in cases where inactive data points abound), their generalization ability is limited. Therefore, these DNN were discarded and not used to generate Snapshot Ensembles. In cases where convergence was indeed achieved, we only kept snapshots with high predictive power; that is, leading to RMSE values on the validation set smaller than 1.2 pIC$_{50}$ units. The training phase of each Snapshot Ensemble stopped once 100 snapshots with a validation RMSE < 1.2 pIC$_{50}$ units were taken. The 100 snapshots were used to predict the activity for both the validation and test sets.

We defined three implementations of this scheme, each of them differing in (i) the number of epochs per cycle, (ii) epoch intervals at which snapshots were taken, and (iii) the step size in the learning rate decay function:

- **Snapshot Ensemble Lr Annealing 1:** a snapshot was taken every 50 epochs, each cycle consisted of 50 epochs, and the learning rate was decreased by 40% every 10 epochs
- **Snapshot Ensemble Lr Annealing 2:** a snapshot was taken every 25 epochs, each cycle consisted of 250 epochs, and the learning rate was decreased by 40% every 10 epochs.
- **Snapshot Ensemble Lr Annealing 3:** a snapshot was taken every 50 epochs, each cycle consisted of 250 epochs, and the learning rate was decreased by 40% every 50 epochs.
- **Random Forests (RF)**

RF models were trained using the python library scikit learn[75]. The default parameter values were used except for the number of trees, which was set to 100 given that higher values do not generally lead to increased performance when modelling bioactivity data[37,76]. Another reason to consider 100 trees was to ensure that the same number of base learners is used to generate RF and DNN ensembles. The same data set splits used to train DNN were used to train RF models. Specifically, RF models were trained on the training set (70% of the data), and the predictive power assessed on the test set (15%).

**Conformal Prediction**

- ***Ensembles of DNN***

We describe below the main steps to generate conformal predictors using ensembles of DNN. For further theoretical details about the conformal prediction framework we refer the reader to [31,32,36].

The residuals and the standard deviation across a given DNN ensemble, be it a Snapshot Ensemble or a DNN Ensemble, served to generate a list of non-conformity values for the validation set as follows (Figure 1B):

$$\text{Equation 1:} \quad \alpha_i = \frac{|y_i - \hat{y}_i|}{e^{\sigma_i}}$$

where $y_i$ is the $i^{th}$ instance in the validation set, $\hat{y}_i$ is the average of the predicted activities for the $i^{th}$ instance across the ensemble, and $\sigma_i$ is the standard deviation of the activity predictions across the ensemble. The resulting list of non-conformity scores, $\alpha$, was sorted in increasing order, and the percentile corresponding to the CL considered was selected, *e.g.*, $\alpha_{80}$ for the 80$^{th}$ percentile. The validation residuals across the 20 runs of each learning strategy and data set combination were used to generate the list of non-conformity values.

Normalizing the residuals is a widely used method to generate tighter predictions for instances that are easier to predict, as reflected by a lower variance across the ensemble[36,77]. We decided to use the natural exponential of the standard deviation across the ensemble to scale the residuals because previous studies showed that this approach slightly improves the efficiency of conformal predictors built on bioactivity data sets[78]. Basically, this scaling sets the upper value for the list of non-conformity values to be equal to the largest residual in the validation set, as the exponential converts low $\sigma_i$ values to ~1. This is useful in practice to prevent large non-conformity values. For instance, a very large non-conformity value would be obtained for a very biased prediction (*i.e.*, large difference between the predicted and the true value), and displaying low variance (*i.e.,* low variability across the base learners). This situation would occur in *e.g.*, the presence of activity cliffs[79].

The standard deviation across the ensemble served to calculate confidence regions for individual instances in the test set as follows:

$$\text{Equation 2:} \quad Confidence\ region = \hat{y}_j \pm |y_j - \hat{y}_j| = \hat{y}_j \pm (e^{\sigma_j} * \alpha_{CL})$$

where $y_j$ is the $j^{th}$ instance in the test set, $\hat{y}_j$ is the average of the predicted activities for the $j^{th}$ instance across the ensemble, and $\sigma_j$ is the standard deviation of the predicted activities across the ensemble.

- *Random Forests*

We also used Random Forests as a baseline comparison, in order to evaluate whether the conformal predictors generated using ensembles of DNNs lead to higher predictive power and tighter confidence intervals than the state of the art. In the case of RF, cross-conformal predictors were generated as previously reported[35,80]. In brief, RF models were trained on the training data using 10-fold cross validation. The cross-validation residuals and the standard deviation across the forest were used to calculate the list of non-conformity values[37,78].

**Results and Discussion**

We firstly evaluated the performance of the different ensemble strategies considered in this study. Overall, the performance of RF, DNN Ensembles, and the three Snapshot Ensembles we defined was comparable on the 24 data sets used in the current study ($P > 0.05$; Kruskal–Wallis test), with mean RMSE values on the test set across repetitions in the 0.6-0.9 pIC$_{50}$ units range (Figure 3). These values are consistent with the expected errors in prediction for models built using heterogeneous IC$_{50}$ data from ChEMBL[56,57]. We furthermore observed a variability of 0.05-0.15 pIC$_{50}$ units across the three Snapshot Ensembles for each data set (Figure 3). The comparable errors in prediction on the test set shows that Snapshot Ensembles performed *on par* with RF and DNN Ensembles. These results thus indicate that the deep learning architectures and parameter values we used to train the four types of DNN ensembles explored in this study are suitable to model these data sets.

To examine the variability of the predictions across snapshots, we computed the correlation between the predictions on the test set for each pair of snapshots in the Snapshot Ensembles, and for the independently trained DNNs that form the Ensemble DNN (Figure 4 and S1-23). The test set predictions calculated by most of the DNN are less correlated in the Ensemble DNN (Figure 4A) than in the Snapshot Ensembles (Figure 4B-D). This is expected because the base learners in the Ensemble DNN are trained independently, whereas the snapshots are taken throughout the training phase of a single neural network. However, we note that the lowest Pearson correlation observed for the Ensemble DNN predictions ( *i.e.,* ~0.92; Figure 3A; see Figures S1-23 for the remaining data sets) is also observed for the Snapshot Ensembles. In the case of Snapshot Ensembles, we also observe that the predictions on the test set are less correlated to the first annealing cycle (*i.e.,* first snapshot) as the training phase advances. This is indicated in Figure 4B-D by the transition from green to brown across the *x* and *y*-axes (see also Figures S1-23). This indicates that during the training phase of Snapshot Ensembles multiple local minima are explored, each of them exhibiting different error profiles, thus helping to increase the diversity in the final ensemble. The predictions for the Snapshots taken throughout each learning rate annealing cycle (*i.e*, Snapshot Ensemble Lr Annealing 2-3), are more correlated than those extracted from other learning cycles, as evidenced by the dark green squares in the diagonal of Figure 4C-D. Note that this pattern is not observed for *Snapshot Ensemble Lr Annealing 1* (Figure 4A), where each snapshot was taken at the end of each annealing cycle. Thus, the high predictive power of each snapshot on the test set, as well as the variability of the predictions across the ensembles, suggest that the snapshots in the Snapshot Ensembles converged to multiple high-quality local minima[81].

Next, we sought to investigate the validity of the conformal predictors generated using DNN Ensembles and Snapshot Ensembles. To this end, we computed the percentage of instances in the test set whose true values lie within the predicted confidence regions. For a conformal predictor to be valid, this fraction needs to be equal to or greater than the selected confidence level. The conformal predictors are valid for all data sets and learning strategies explored, as indicated by the high correlation between increasingly higher confidence levels and the percentage of confidence intervals encompassing the true bioactivity value ($R^2 > 0.99$, $P < 0.001$; Figures 5 and S24). Together, these results indicate that the four types of DNN-based ensembles considered here can be used to generate valid conformal predictors.

The second critical aspect of a conformal predictor is the *efficiency* of the predicted confidence regions. Following the terminology used in the conformal prediction literature[36], we use the term efficiency to refer to the average size of the predicted confidence intervals across models. The narrower the confidence intervals, the more efficient a conformal predictor is. Conformal predictors generating large intervals might be theoretically valid, but such wide errors bars would provide little practical information about which compounds to prioritize for further experimental testing in a real-world situation. Thus, we next compared the size of the confidence regions predicted by the conformal predictors based on RF, DNN Ensembles, and Snapshot Ensembles.

The average size of the confidence regions is overall comparable across data sets for RF and DNN Ensemble models, and slightly higher for Snapshot Ensembles (Figure 6). The average size over the 20 repetitions of the confidence intervals is in the 0.8-1.2 $pIC_{50}$ range for all learning strategies and data sets. These values are comparable to those previously obtained for other bioactivity data sets[36,78], and are narrow enough to be practically useful for compound prioritization in most cases. Notably, the RF-based conformal predictors span a wider range of values, leading to confidence intervals larger than 6 $pIC_{50}$ units in some cases (Figure 6). Although the spread of the distribution is larger for those Snapshot Ensembles with lower predictive power on the test set (Figure 3), at least one Snapshot Ensemble leads to a tighter distribution for all data sets (Figure 6). This indicates that, although robust conformal predictors can be generated using Deep Confidence, the parameterization of the underlying networks needs to be optimized for each individual data set.

The fact that the confidence intervals calculated with RF span a wider range of values (Figure 6) indicates that the predictions of the individual trees in the Forest are poorly correlated for some instances (*i.e.*, high variance across the ensemble), and highly correlated for others (*i.e.*, low variance). High variance across the ensemble leads to larger confidence intervals (Equation 2). In contrast, the confidence intervals for molecules for which the predictions across base learners are highly correlated, and hence display a lower variance, are tighter. Thus, lower variance across the ensemble should be reflected in more narrowly contained distributions of intervals sizes[82]. Analysis of the distributions of the standard deviation across the ensemble for the predictions on the test set confirms that the RF models display larger variance than DNN-based ensembles, with mean values for RF around 3 times larger (Figure 7), thus explaining the differences observed in Figure 6.

The validity of conformal predictors is guaranteed for the entire set of predictions, *i.e.,* global validity[36]. However, the local validity, *e.g.,* the validity of a conformal predictor if measured using data within a certain bioactivity range only, might not always be fulfilled. This is particularly important because the goal of most discovery projects is to find highly active compounds, and hence, guaranteeing that the error rates are correct for particular bioactivity ranges is of utmost importance. To further examine the local validity of the DNN-based conformal predictors, we calculated the error rate at confidence level 0.8 for each $pIC_{50}$ unit bin, using the experimentally measured $pIC_{50}$ values for binning the instances in the test set. We find that the error rate is around the expected value (0.20 in this case; *i.e.,* 1 - 0.8) for most conformal predictors and data sets in the low micromolar and high nanomolar ranges (*i.e.,* $pIC_{50}$ of 5-7), although notable exceptions are present, *e.g.*, data sets Caspase, COX-2, or HERG (Figures 8-9 and S25-26). Although not consistently, RF often shows a higher error rate for particular $pIC_{50}$ bins (see *e.g.,* data sets Ephrin and erbB1 in Figure 9). We note that the error rates are higher for regions with fewer instances, which correspond for most data sets to the high activity range. Future work will be needed to develop methods to guarantee the local validity for the high activity range, which is usually the range of interest and the one with fewer training instances. Together, these results indicate that the errors measured for DDN-based conformal predictors are more even than for RF-based ones, as evidenced by the lower spread of the distributions of confidence intervals and, in some cases, the more evenly distributed error rates across $pIC_{50}$ bins.

Overall, we have shown here that it is possible to generate highly predictive models and valid conformal predictors using DNN Snapshot Ensembles. Of practical relevance to predictive modelling in drug discovery[38], the derived confidence intervals display a smaller spread than that of RF-based conformal predictors. The performance for ensembles of independently trained DNN and Snapshot Ensembles is comparable. Thus, the framework proposed here is computationally efficient in that it just requires the training of a single neural network to generate confidence intervals. The parameter values used were found to work well on these data sets, given that model performance was in line with previously published analyses. However, further tuning of the deep learning architecture or the parameters would be advised for other learning tasks, data sets or compound descriptors[13,18].

We also stress that the potential of deep learning has not been fully exploited in this work due to the limited size of the data sets used. Although the performance of DNN ensembles and RF on these data sets is comparable, DNN will likely outperform RF on larger data sets[23]. It is precisely for the modelling of large data sets that the framework proposed here will prove more versatile to leverage both the increased predictive power of deep learning and the estimation of prediction reliability using conformal prediction, while requiring minimal extra computation. While the generation of snapshots during the training of a DNN requires no extra computation, we stress the fact that additional computation might be required to generate a Snapshot Ensemble if the network converges to high-quality local minimum before the number of epochs required to generate 100 snapshots is reached. This is however unlikely given that the global minimum is almost never reached in practice for deep architectures. We also note that the complexity of the loss landscape increases for deeper architectures[83], and hence, sampling multiple local minima to generate a Snapshot Ensemble might require a more sophisticated learning rate annealing schedule in those cases. Future work will also be required to comprehensively evaluate whether the exhaustive sampling of local minima in the loss landscape leads to more efficient conformal predictors, as opposed to the sampling of a reduced set of these generating correlated predictions, or just sampling the vicinity of few local minima[81].

**Conclusions**

In this work we introduced Deep Confidence, a framework to compute valid and efficient confidence intervals for deep neural networks that requires no extra computational cost. Specifically, we have shown that snapshots taken throughout the training of a single neural network and corresponding to different local minima provide sufficient information to predict confidence regions that are narrower than those calculated with RF-based conformal predictors. Therefore, the framework proposed in this study has great potential to increase the application of deep neural architectures in early-stage drug discovery while also controlling for the reliability of individual predictions.

**Figures**

**Figure 1 Deep Confidence workflow.** Illustration of the generation of a Snapshot Ensemble from the training of a single deep neural network (A), and the workflow proposed in this study to generate conformal predictors from Snapshot Ensembles (B).

**Figure 2 Loss on the validation set during the training of DNN using cyclical learning rate annealing.** Evolution of the loss (shown in black, and corresponding to the RMSE for the observed against the predicted values for the validation set), best loss for each learning rate annealing cycle of 250 epochs (red), and the learning rate (shown in blue and annealed using step decay) for three representative data sets and across the first 1,500 epochs of the training phase. This learning rate annealing scheme corresponds to the learning strategy referred to in the text as *Snapshot Ensemble Adjust Lr 3*. It can be seen that the loss converges to a local minimum during each learning rate annealing cycle of 250 epochs, throughout which the learning rate is decreased by 40% every 50 epochs. It can be seen in (C) that the loss on the validation set decreases after multiple learning rate annealing cycles.

**Figure 3 Performance on the test set.** Mean RMSE values (+/- standard deviation) on the test set across 20 runs for RF, DNN Ensembles, and Snapshot Ensembles are shown. Overall, the three types of Snapshot Ensembles considered showed high predictive power on the test data, indicating that the choice of parameter values is suitable for modelling these data sets. The color scheme used in this Figure is also used throughout subsequent figures in the manuscript.

**Figure 4 Analysis of the correlation of the base learner predictions on the test set for Ensemble DNN and Snapshot Ensembles (data set A2a).** The pairwise correlation of the test set predictions was calculated for each pair of base learners forming the Ensemble DNN (A) and the three types of Snapshot Ensembles we defined (B-C). The correlation matrices were calculated using the predictions on the test set for the 20 repetitions. Overall, the test set predictions calculated by each DNN are less correlated in the Ensemble DNN than in the Snapshot Ensembles. This is expected given that the base DNN in the Ensemble DNN are independently trained. The dark green squares in the diagonal in C and D indicate that the predictions calculated with snapshots extracted from the same learning rate annealing cycle are more correlated than those extracted from other learning cycles. The transition from green to brown across the training phase observed for Snapshot Ensembles indicates that multiple local minima are explored in each annealing cycle throughout the training phase, and hence, the predictions on the test set are less correlated to the first annealing cycle as the training phase advances. A similar analysis for the other data sets is provided in Figures S1-23.

**Figure 5 Validity analysis.** Each plot shows the correlation between the confidence level and the percentage of predictions whose true value is contained within the predicted intervals for the conformal predictors built on 4 representative data sets using RF (A), DNN Ensembles (B), and Snapshot Ensembles Adjust Lr 1-3 (C-E). It can be seen that valid models were obtained because the percentage of predictions whose true value lies within the predicted confidence region (1- error rate) strongly correlates with the confidence level. The validity analysis for the remaining data sets is provided in Figure S24.

**Figure 6 Analysis of the distribution of the confidence intervals**. Each box plot shows the distribution of the size of the confidence intervals (in $pIC_{50}$ units) calculated for the test set at confidence level 80%. The predictions for the 20 runs are shown. Overall, it can be seen that the spread of the distributions for DNN-based methods is smaller, while the average size for the confidence regions is marginally higher, compared to RF.

**Figure 7  Ensemble variance analysis**. Distributions of the standard deviation across the ensemble for the test set instances are shown across the 20 runs. Overall, it can be seen that the variance across the ensemble is significantly larger for RF as compared to DNN-based ensembles. This affects the spread of the distribution of intervals sizes, as shown in Figure 5.

**Figure 8  Analysis of the error rate across pIC$_{50}$ bins**. The error rates calculated for those instances in the test set with an observed pIC$_{50}$ value in the range indicated in the *x*-axis at confidence level 80% are shown. The horizontal line indicates the expected error rate. According to the expected error rate, all points should be below the line. Points over the horizontal line indicate that the error rate is higher that the indicated by the chosen confidence level. The area of the dot shapes is proportional to the number of instances in each pIC$_{50}$ bin. Overall, it can be seen that even if the global validity is kept (*i.e.,* the error rate for all test set instances corresponds to the expected error rate for the selected confidence level; see Figure 5), the validity across pIC$_{50}$ bins varies considerably. The same analysis for the other data sets is provided in Figure 9.

**Figure 9  Analysis of the error rate across pIC$_{50}$ bins for the remaining data sets**. The error rates calculated for those instances in the test set with an observed pIC$_{50}$ value in the range indicated in the *x*-axis at confidence level 80% are shown. The horizontal line indicates the expected error rate. See also Figure 8.

## Author Contributions

I.C.-C. designed research, trained the models, and analyzed the results. I.C.-C. generated the figures and wrote the paper with substantial input from A.B.

## Acknowledgements

This project has received funding from the European Union's Framework Programme For Research and Innovation Horizon 2020 (2014-2020) under the Marie Curie Sklodowska-Curie Grant Agreement No. 703543 (I.C.C.).

## Conflicts of Interest

The authors declare no conflict of interests.

**Supporting Information Available:** The Supporting Information consists of (i) Figures S1-23 reporting the correlation of the predictions on the test set for each pair of base learners in the Ensemble DNN and the three types of Snapshot Ensembles explored, (ii) Figure S24 reporting the validity of the conformal predictors built for the 21 data sets not shown in Figure 4, (iii) Figures S25-26 reporting the validity of the conformal predictors across $pIC_{50}$ bins calculated using the predicted $pIC_{50}$ values, and (iv) the 24 data sets used in this study.

**Figure 1**

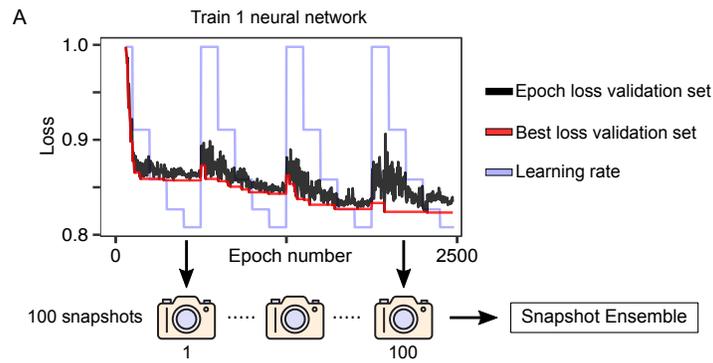

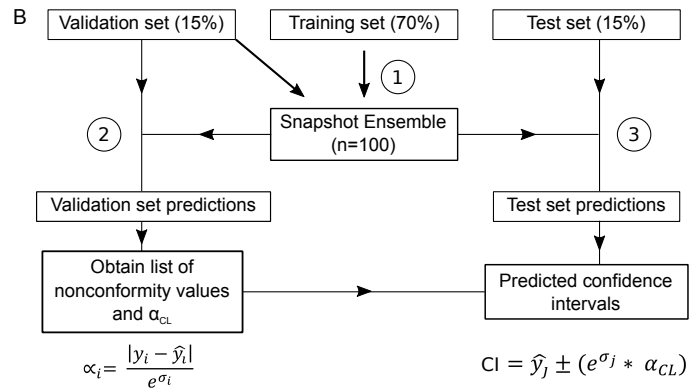

**Figure 2**

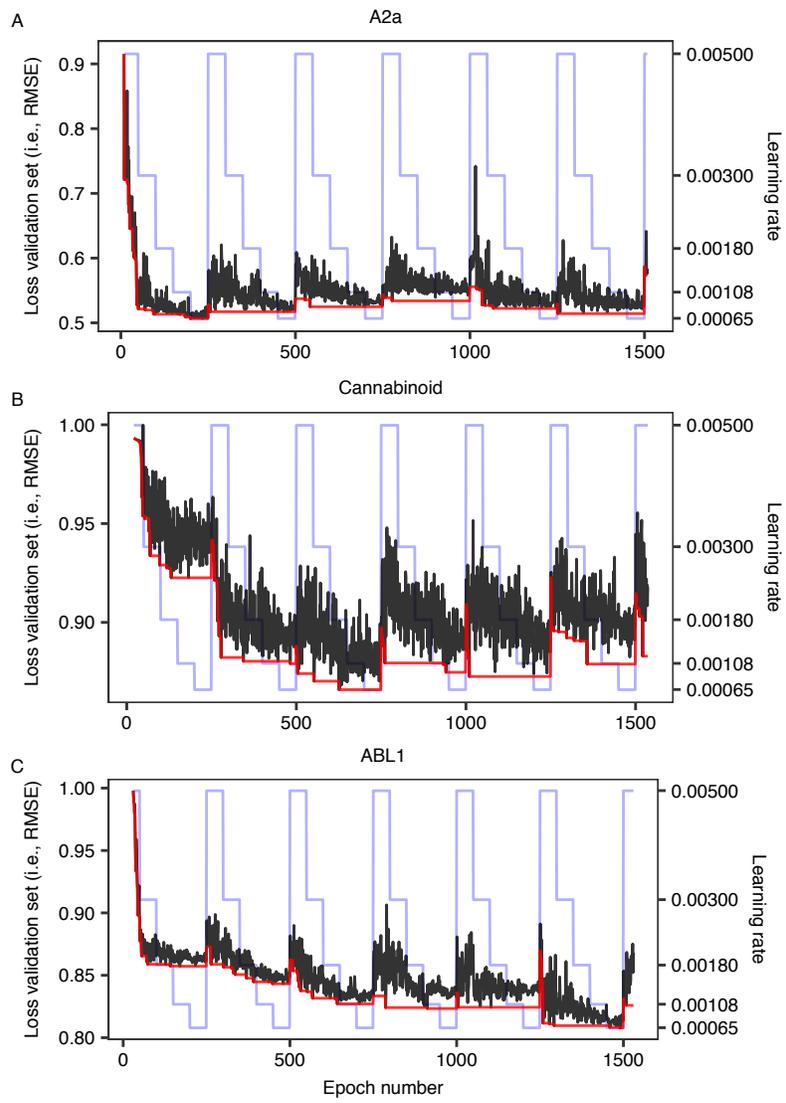

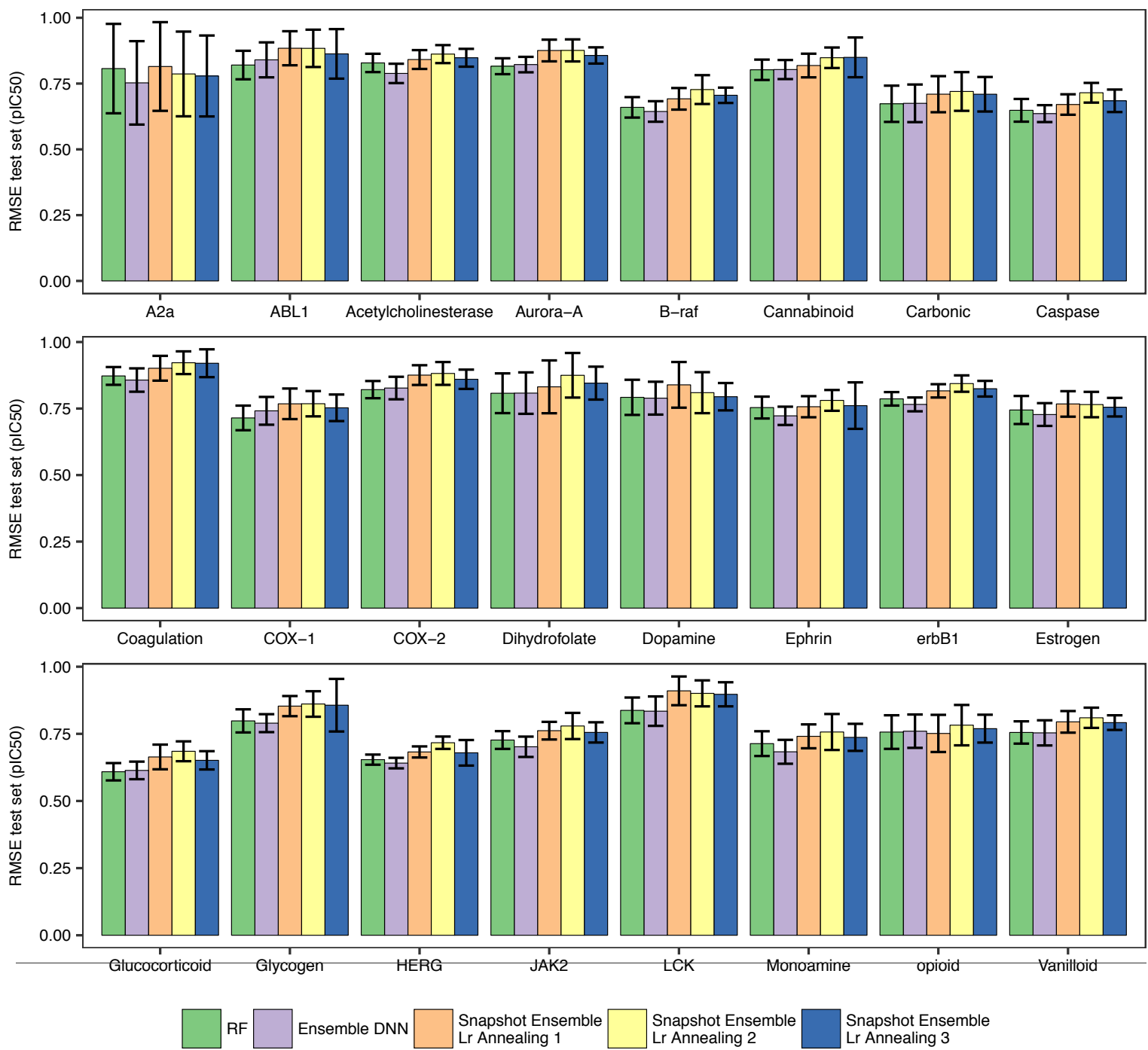

**Figure 3**

**Figure 4**

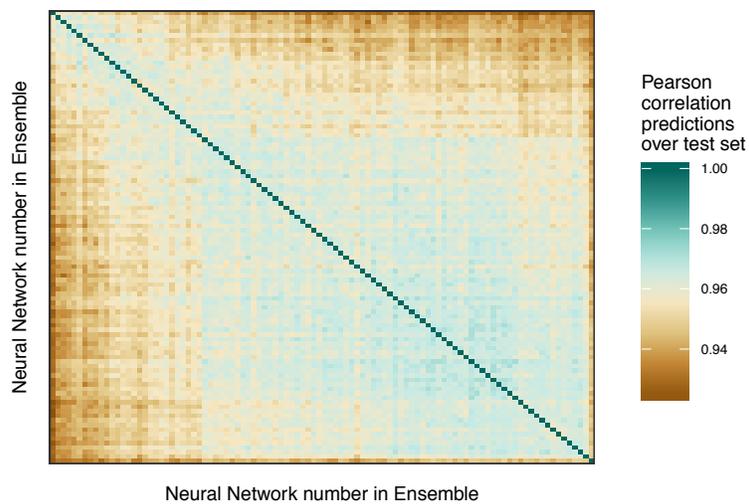

A  Ensemble DNN

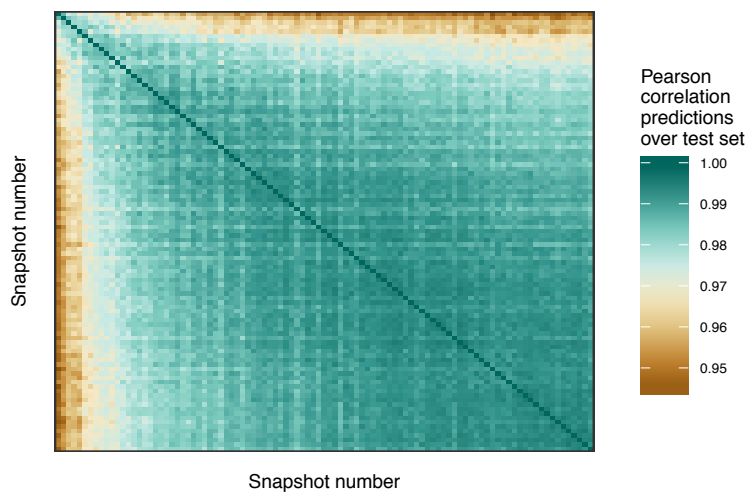

B  Snapshot Ensemble Lr Annealing 1

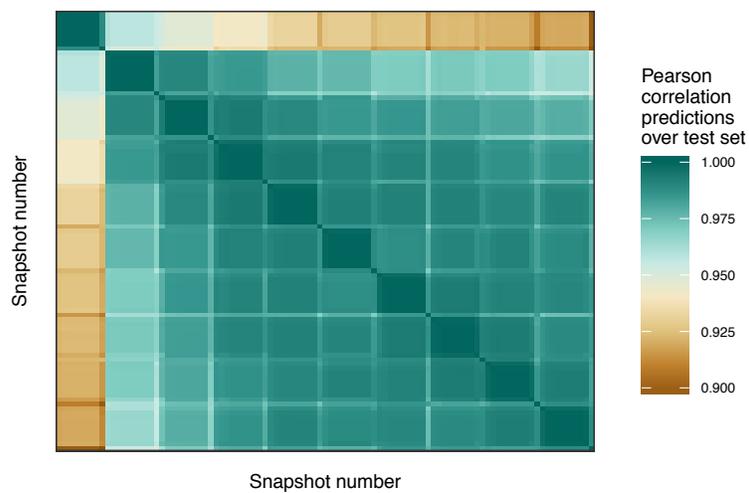

C  Snapshot Ensemble Lr Annealing 2

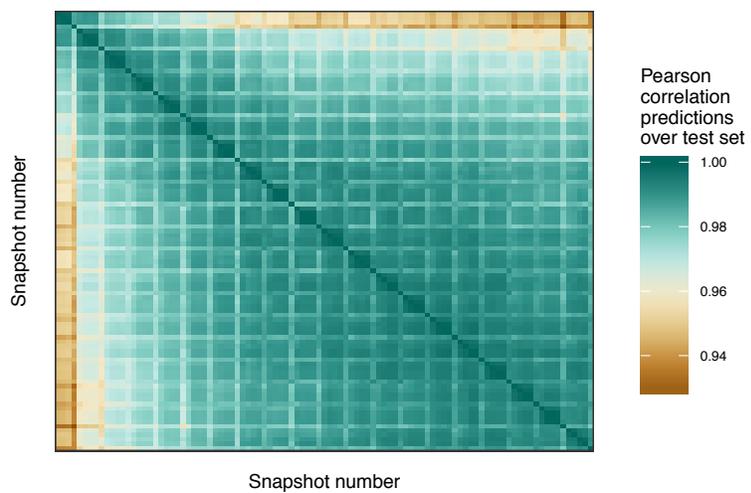

D  Snapshot Ensemble Lr Annealing 3

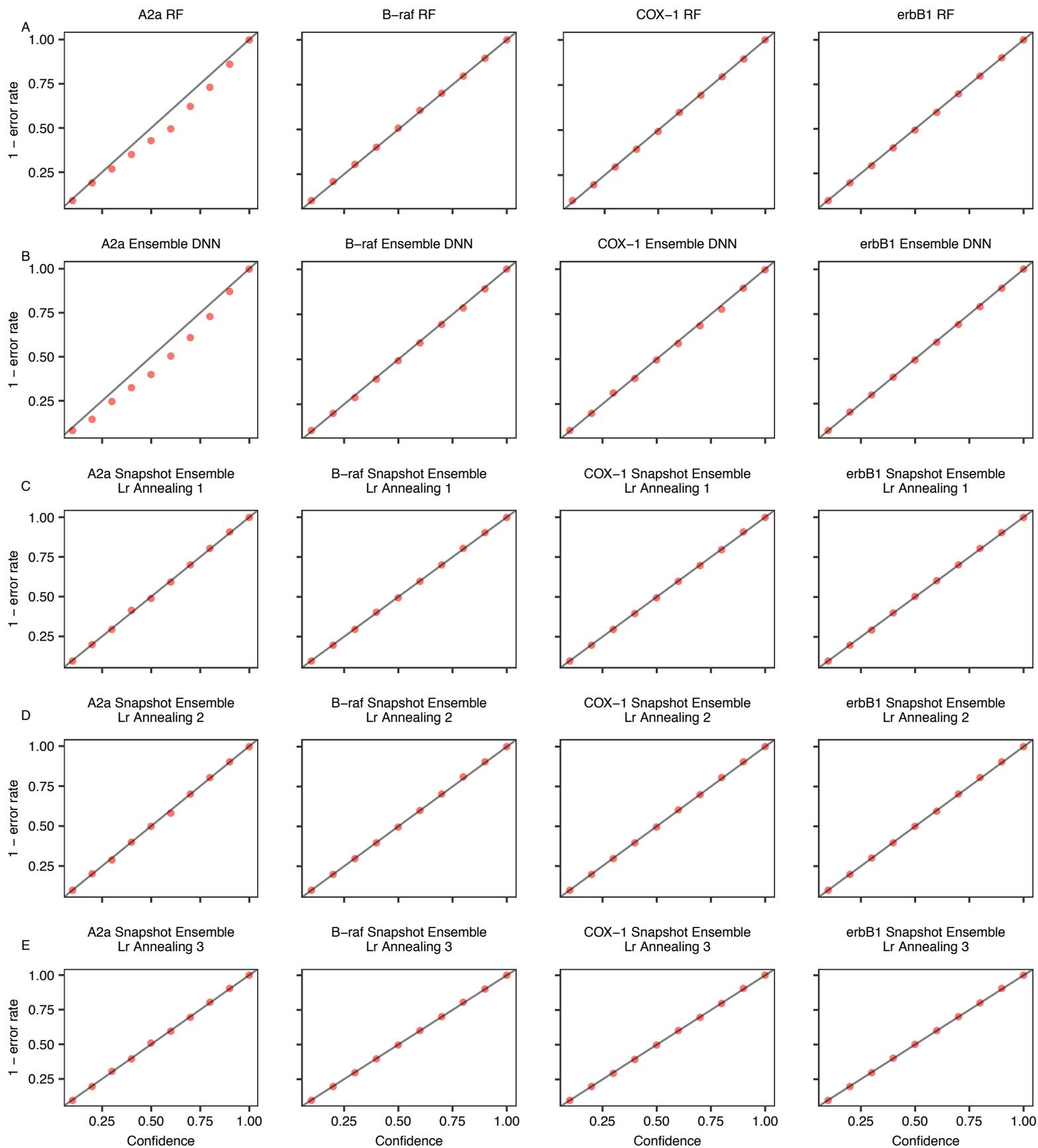

Figure 5

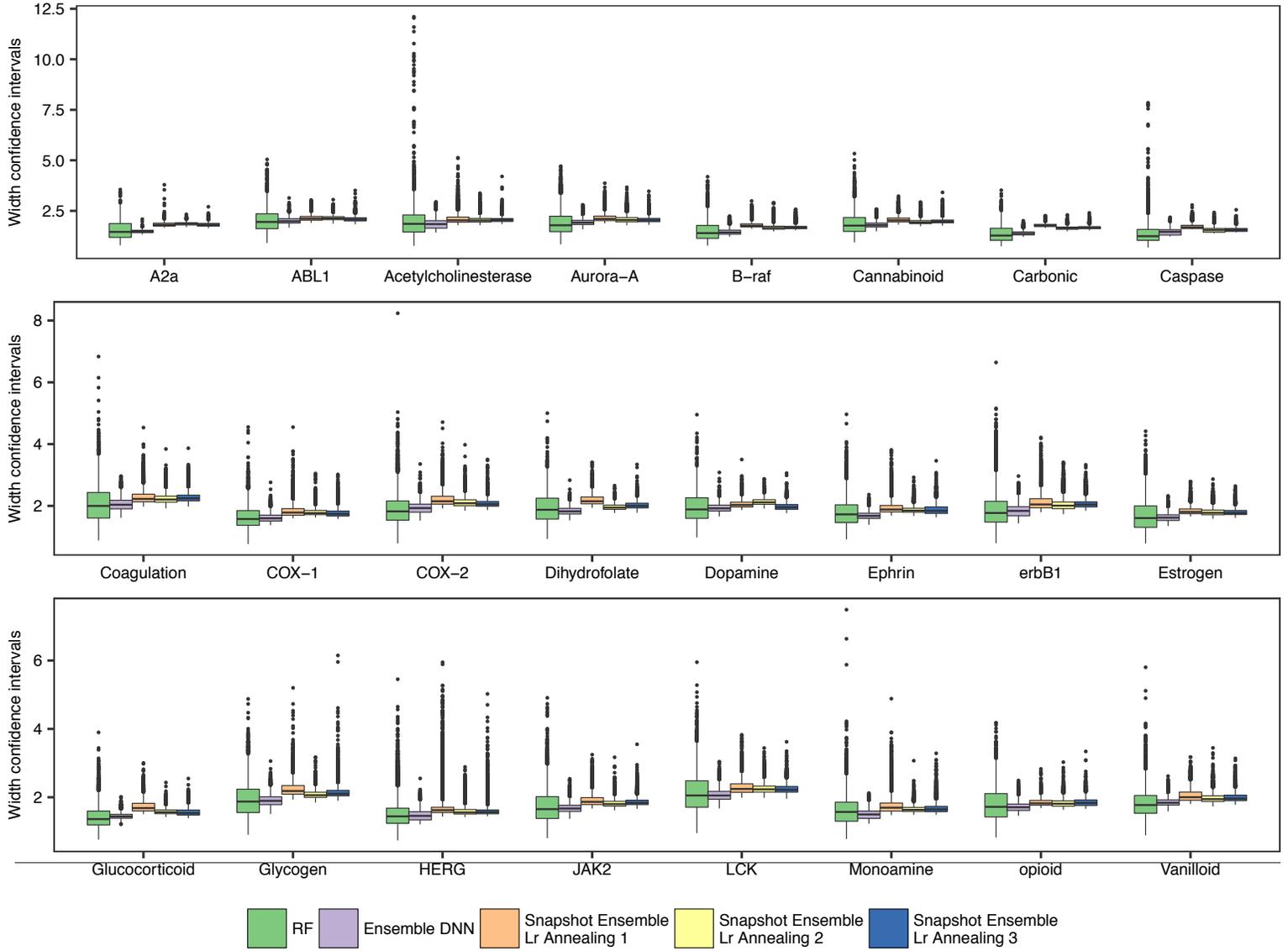

# Figure 7

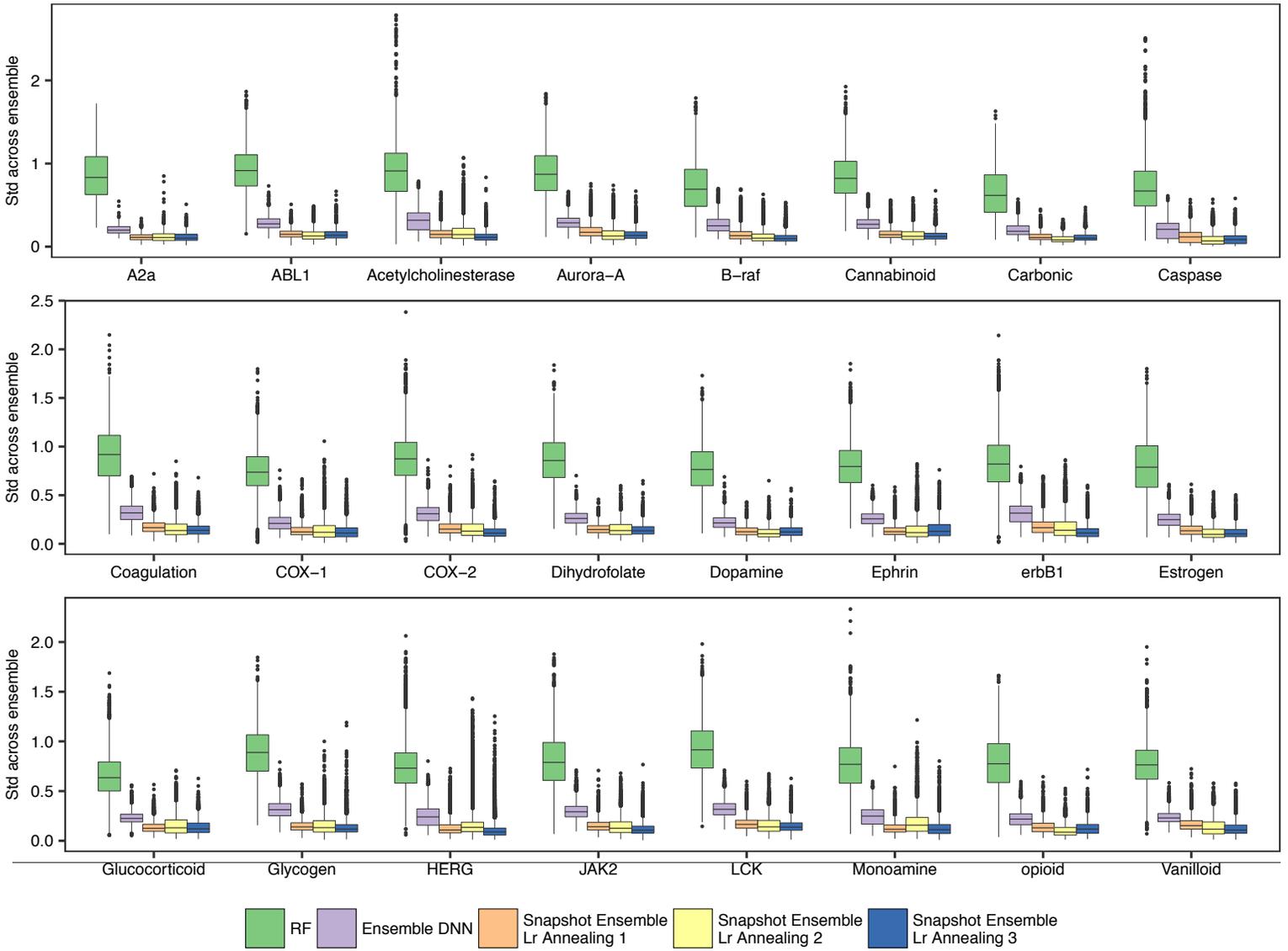

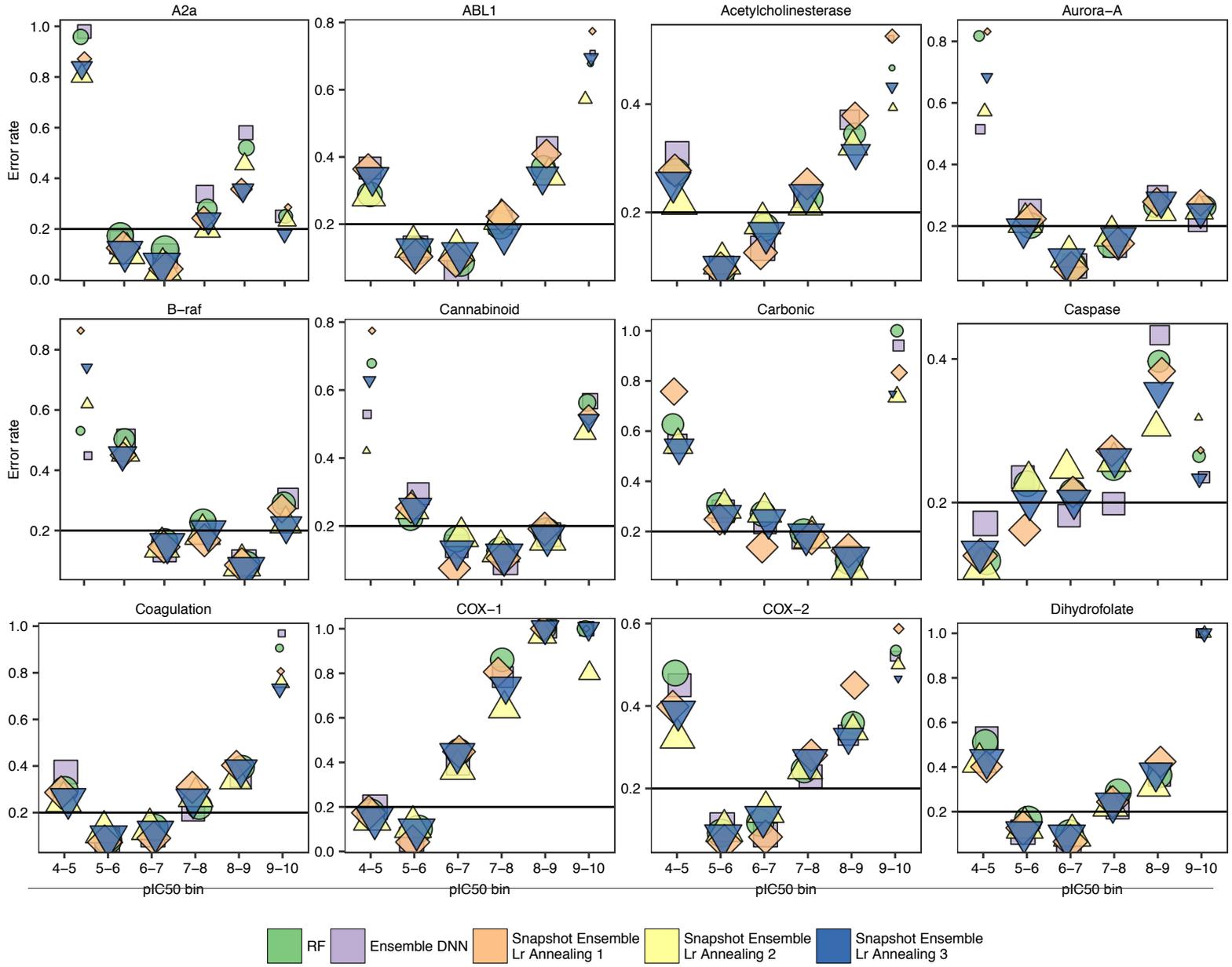

Figure 8

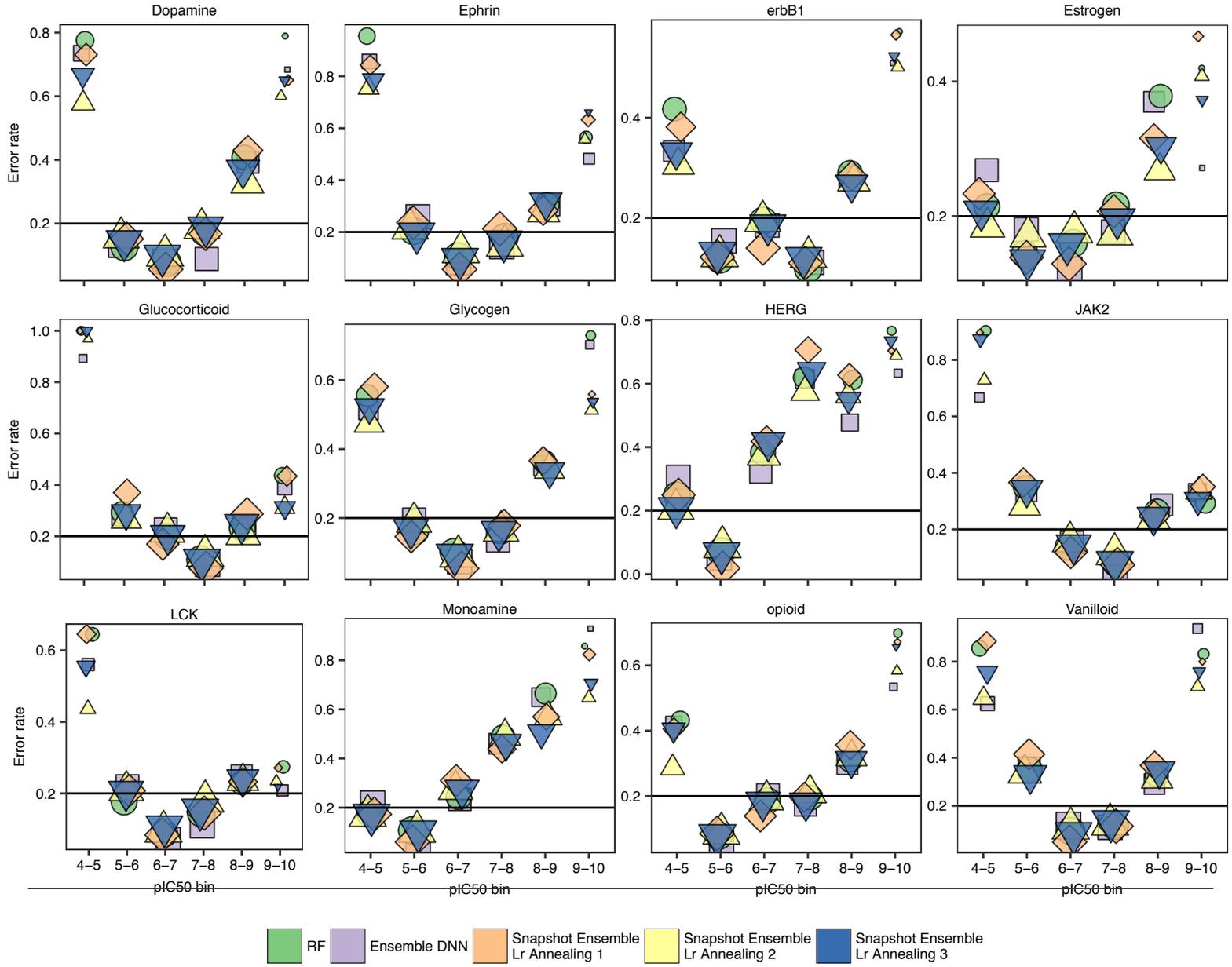

Figure 9